\documentclass[11pt]{article}

\usepackage[margin=1in]{geometry}

\usepackage{amsthm,amsmath,amssymb,enumitem}
\usepackage{authblk,setspace,booktabs,graphicx}
\usepackage[numbers,sort&compress]{natbib}
\usepackage{xcolor,multirow,booktabs}
\usepackage{lineno,comment,float}
\usepackage{hyperref}

\usepackage{algorithm,algpseudocode,algorithmicx}
\algrenewcommand\algorithmicrequire{\textbf{Input:}}
\newcommand{\Input}{\Require}

\theoremstyle{plain}
\newtheorem{theorem}{Theorem}[section]
\newtheorem{remark}[theorem]{Remark}

\newtheorem*{lemma*}{Lemma}
\newtheorem*{theorem*}{Theorem}
\newtheorem*{claim*}{Claim}
\newtheorem*{remark*}{Remark}

\def\X{{\mathcal{X}}}

\author[1]{Shayan Shafaei\thanks{Email: shayan.shafaei@ou.edu}}
\author[1]{Naeem Shahabi Sani}
\author[1]{Chao Lan}
\affil[1]{School of Computer Science, 
University of Oklahoma}
\date{}

\begin{document}

\title{Randomized Weighted Distance 
Nearest Neighbor Classification}

\maketitle

\begin{abstract}
The nearest neighbor classifier is 
known to be memory-inefficient, 
and a popular solution is to 
compress its training data. 
Recently, a compression algorithm 
based on weighted distance nearest neighbor 
(WNN) was proposed, which has elegant  
theoretical properties and effective 
classification performance. However, 
its compression process is computationally expensive, 
for it performs a greedy search of the 
densest homogeneous balls in 
data space. In this paper, we propose 
a randomized variant RWNN. It 
simply randomizes the ball search process, 
and results in substantially reduced 
compression time (often 0.1\%--1\% of WNN's) 
and similar classification accuracy. 
RWNN also inherits several theoretical 
properties of WNN including consistency 
and a generalization bound. Experimental 
results on multiple datasets demonstrate 
its efficacy.
\end{abstract}

\section{Introduction}

\label{sec:intro}

The nearest neighbor classifier has been long 
studied for its simplicity and interpretability \cite{CoverHart1967,chen2018explaining}, but 
it is also known to be memory-inefficient 
due to the need of storing all training data 
for inference. A popular solution is to 
compress the training set, but the compression 
process may be expensive and thus lead to a hard 
tradeoff between memory and computation efficiency. 

An example is the recent compression algorithm 
weighted distance nearest neighbor (WNN) \cite{Gottlieb2024}. 
It is theoretically elegant and empirically 
effective, but the compression process is 
computationally expensive as it relies on 
a greedy search for the densest homogeneous 
balls in data space. Here, a homogeneous ball 
is one that contains data from the same class. 
Details will be provided later. 

This research leverages randomness to 
accelerate WNN. 
We show that, by simply randomizing 
the ball search process (i.e., randomly 
pick a homogeneous ball instead of  
finding the densest one), 
the algorithm can be significantly 
accelerated without sacrificing empirical 
effectiveness or certain theoretical properties. 
We call this new variant randomized WNN (RWNN), 
and further extend it to a heuristic RWNNs which 
terminates the random-pick process early to achieve 
desired compression rate. 
Experimental results on multiple datasets 
demonstrate efficacy of our algorithms, where they 
often require less than 1\% of the WNN time while maintaining classification accuracy. 

\subsection{Related Work}

 Data compression for nearest neighbor classification has been long studied. 
 Early representative methods 
 include condensed nearest neighbor (CNN) \cite{Hart1968CNN}, 
 reduced nearest neighbor (RNN) 
 \cite{Gates1972}, and fast condensed nearest neighbor (FCNN) \cite{Angiulli2005}. 
 Their main idea is to have the compressed 
 set repeatedly include training points that are 
 misclassified by it. Modern studies 
 explore other directions, e.g.,  
near-optimal sample compression (NOSC) \cite{gottlieb2014near} is 
the first method with provable guarantees 
on compression rate and generalization 
performance, and stochastic neighbor compression (SNC) \cite{kusner2014stochastic} generates 
a synthetic compressed set which often 
achieves higher compression rates. 
More recently, weighted distance nearest 
neighbor condensing (WNN) \cite{Gottlieb2024} 
aims to find a compressed set that is 
optimal for a distance-weighted nearest 
neighbor classifier. The resulting classifier 
is shown to enjoy many theoretical properties, 
including consistency (i.e., correctly 
classify all training points), Bayes 
consistency and generalization. 
In this paper, we improve the 
computational efficiency of WNN.

\section{Preliminaries}

Following \cite{Gottlieb2024}, 
we focus on binary classification. 
Let $\mathcal{X} \subseteq \mathbb{R}^p$ 
be the data space,  $d:\mathcal{X}\times\mathcal{X}\to\mathbb{R}$ 
be a distance function, and 
$\ell:\mathcal{X}\to\{0,1\}$ 
be the labeling function, i.e., 
$\ell(x)$ is the true label of $x$.

Given a training set $S\subseteq \mathcal{X}$, 
the nearest neighbor classifier is 
\begin{equation}
h(z)=\ell\!\left(\arg\min_{x\in S} d(z,x)\right), 
\quad \forall z \in \mathcal{X}.
\end{equation}

The WNN classifier \cite{Gottlieb2024} 
is built on ``homogeneous ball''. 
For an $x\in\X$ and $r>0$, let 
\begin{equation}
B(x,r)=\{z\in\X: d(z,x)<r\}, 
\end{equation}
be the open ball centered at $x$ with radius $r$. 
A homogeneous ball centered at $x$ is the largest open ball that only contains $S$ points which have 
the same class as $x$, and the ball radius 
is the distance from $x$ to the nearest 
$S$ point from a different class, i.e.,  
\begin{equation}
\label{eq:rset}
r(x) = \min_{z\in S:\,\ell(z)\neq \ell(x)} d(x,z).
\end{equation}

Given a compressed set $\tilde{S}\subseteq S$, 
the WNN classifier is then defined as 
\begin{equation}
\label{eq:wnnclassifier}
h_{(\tilde{S},w)}(z) =
\ell\!\left(\arg\min_{x\in\tilde{S}} 
d_w(z,x)\right),  
\end{equation}
where $d_w: \mathcal{X} \times 
\mathcal{X} \to \mathbb{R}$ is 
a weighted distance function defined as 
\begin{equation}
\label{eq:distanceweight}
d_w(x,x') = \frac{d(x,x')}{w(x)\,w(x')},
\end{equation}
and $w:\X\to\mathbb{R}$ is a weighting function 
which assigns $1$ to all points not in $\tilde{S}$ 
and assigns the radii of homogeneous balls 
centered at points in $\tilde{S}$. 

Let $\mathrm{err}(h)=\Pr(h(x)\neq \ell(x))$ be 
the error of $h$. 
The following is from 
\cite[Theorem 3.4]{Gottlieb2024}.

\begin{theorem}[\cite{Gottlieb2024}]
\label{thm:error}
For any distribution on $\mathcal{X}$, labeling function $\ell$, and $\delta\in(0,1)$, with probability at least $1-\delta$ over the i.i.d. sampling of $S$, if the WNN classifier $h_{(\tilde{S},w)}$ corresponding to $\tilde{S}$ and $w$ correctly classifies all points in $S$, then
\begin{equation}
\label{eq:rwnn-bound}
\mathrm{err}(h_{(\tilde{S},w)})
\leq
\frac{2}{|S|-|\tilde{S}|}
\left(
|\tilde{S}|\ln\frac{2e|S|}{|\tilde{S}|}
+ \ln\frac{|S|}{\delta}\right).
\end{equation}
\end{theorem}

It is verified in \cite{Gottlieb2024} 
that the WNN classifier $h_{(\tilde{S},w)}$ 
indeed correctly classifies all points in $S$ 
(i.e. being consistent), 
and thus enjoys the above generalization bound.

\section{Randomized WNN}

In this section, we introduce randomized weighted 
distance nearest neighbor condensing (RWNN). Our
compression algorithm is almost the same as WNN, except that it randomly picks homogeneous balls instead of finding the densest ones. 
Detailed process is presented in Algorithm \ref{alg:rwnn}. 

\begin{algorithm}[t!]
\caption{RWNN Compression}
\label{alg:rwnn}
\setstretch{1.1}
\begin{algorithmic}[1]
\vspace{5pt}
\Input Training set $S$
\State Candidate set $S' \gets S$.
\State Condensed set $\tilde{S} \gets \emptyset$. 
\State Weight set $w$, where 
$w(x) \gets 1$ for all $x \in S$.
\While{$S' \neq \emptyset$}
  \State Randomly pick an $x \in S'$.
  \State Compute radius of the homogeneous ball 
  centered at $x$ as $r(x) = \min_{z\in S:\,\ell(z)\neq \ell(x)} d(x,z)$. 
  \State Add $x$ to the condensed set, i.e., 
  $\tilde{S} \gets \tilde{S} \cup \{x\}$. 
  \State Update the weight of $x$ 
  to the ball radius, 
  i.e., $w(x) \gets r(x)$.
  \State Remove all points in the ball, i.e.,
  $S' \gets S' - \{z \in S' \cap B(x,r(x)) \}$.
  \EndWhile
\State 
\Return $\tilde{S}$ and $w$.
\end{algorithmic}
\end{algorithm}

Our main innovation lies in Steps 5-6, 
which randomly pick $x$ and then compute 
its ball radius -- this takes $O(|S|)$ 
time per round. 
In contrast, WNN needs to compute 
the ball radius for every $x \in S$, 
which takes $O(|S|^2)$ time; 
then, it repeatedly finds the densest balls 
which takes $\Omega(|S'|)$ time per round. 
This is the main reason for our speedup. 

After the condensed set $\tilde{S}$ 
and weight set $w$ are returned, 
we apply the same classifier as WNN, 
defined in (\ref{eq:wnnclassifier}). 
For clarity, we will refer to the 
resulting classifier as the RWNN classifier.

\subsection{Theoretical Properties}

RWNN inherits two theoretical properties of WNN, 
as these properties only depend on the homogeneity 
of balls and classifier design but not the density of balls. The first property is consistency. 

\begin{remark}
\label{rem:rwnn}
The RWNN classifier returned by Algorithm \ref{alg:rwnn} is consistent on the input 
training set, i.e., it can correctly classify 
all points in $S$.
\end{remark}

\begin{proof}
The argument is the same as in \cite{Gottlieb2024}. 
Let $h_{(\tilde{S}, w)}$ be the RWNN classifier, 
where $\tilde{S}$ and $w$ are the condensed set 
and weight set returned by 
Algorithm \ref{alg:rwnn}. Consider two cases. 

\vspace{5pt} \noindent 
\textit{Case 1}: $z \in \tilde{S}$. 
Then, it will be correctly classified
by $h_{(\tilde{S}, w)}$ as $z$ is 
its own nearest neighbor. 

\vspace{5pt} \noindent 
\textit{Case 2}: $z \not\in \tilde{S}$. 
In this case, $z \in B(x,r(x))$ for some 
$x \in \tilde{S}$ (and is removed with 
the ball at Step 9). By the definition of 
$r(x)$, this means $\ell(z) = \ell(x)$. 
In addition, $w(z) = 1$ and $w(x) = r(x)$
by Steps 3 \& 8. 
Then, by (\ref{eq:distanceweight}), 
the weighted distance from $z$ to $x$ is
\begin{equation}
\label{eq:consistency01}
d_w(z,x) = \frac{d(z,x)}{w(z)w(x)} 
= \frac{d(z,x)}{1 \cdot r(x)} <1, 
\end{equation}
where the inequality holds because 
$z \in B(x,r(x))$. 

Now let $x' \in \tilde{S}$ be 
any point with a different label 
from $x$. Then 
\begin{equation}
\label{eq:consistency02}
d_w(z,x') = 
\frac{d(z,x')}{w(z) w(x')}
= \frac{d(z,x')}{1 \cdot r(x')}
\geq 1,     
\end{equation}
where the inequality holds because 
$x'$ is from a different class 
so $d(z,x') \geq r(x')$ 
by definition (\ref{eq:rset}). 

Combining (\ref{eq:consistency01}, 
\ref{eq:consistency02}) shows 
that no point from a different class 
can be the nearest neighbor of $z$, 
because their weighted distance to $z$ is 
at least 1 and strictly bigger than 
$d_w(x,z) < 1$ for the ball center $x$ 
which satisfies $\ell(x) = \ell(z)$. Thus, 
the nearest neighbor rule 
will correctly classify $z$. 
\end{proof}

The second property is that the RWNN 
classifier inherits the generalization error 
bound of the WNN classifier, because it is 
also consistent on the training set 
(Remark \ref{rem:rwnn}). 

\begin{remark}
\label{rem:error}
Theorem~\ref{thm:error} applies to the RWNN classifier built on Algorithm~\ref{alg:rwnn}.
\end{remark}

The only difference is that, intuitively, 
the condensed set of RWNN is expected to be 
larger than 
the condensed set of WNN because the latter 
removes more points through finding the 
densest balls. This means WNN results in 
a smaller $|\tilde{S}|$, which implies a 
smaller error bound in Theorem~\ref{thm:error}. 

\subsection{A Heuristic Variant}

In this section, we propose a heuristic 
variant of RWNN (RWNNs), which can further 
control the condensed set size. Its idea 
is to simply stop the homogeneous ball search 
process when a desired condensing rate is 
achieved. This allows RWNNs to achieve 
the same condensing rate as WNN, although 
it also loses the consistency guarantee 
because $S'$ is not exhausted. 
Detailed process is described in 
Algorithm \ref{alg:rwnns}, where 
Step 4 has an additional stopping 
criterion.

\begin{algorithm}[t!]
\caption{RWNNs Compression}
\label{alg:rwnns}
\setstretch{1.1}
\begin{algorithmic}[1]
\vspace{5pt}
\Input Training set $S$, threshold $\theta$.
\State Candidate set $S' \gets S$.
\State Condensed set $\tilde{S} \gets \emptyset$. 
\State Weight set $w$, where 
$w(x) \gets 1$ for all $x \in S$.
\While{$S' \neq \emptyset$ 
and $|\tilde{S}|/|S| < \theta$}
  \State Randomly pick an $x \in S'$.
  \State Compute radius of the homogeneous ball 
  centered at $x$ as $r(x) = \min_{z\in S:\,\ell(z)\neq \ell(x)} d(x,z)$.
  \State Add $x$ to the condensed set, i.e., 
  $\tilde{S} \gets \tilde{S} \cup \{x\}$. 
  \State Update the weight of $x$ 
  to the ball radius, 
  i.e., $w(x) \gets r(x)$.
  \State Remove all points in the ball, i.e.,
  $S' \gets S' - \{z \in S' \cap B(x,r(x)) \}$.
  \EndWhile
\State 
\Return $\tilde{S}$ and $w$.
\end{algorithmic}
\end{algorithm}

\section{Experiment}
\label{sec:exp}

We evaluated the proposed algorithms 
on ten real-world datasets, including 
Aids \cite{tanshihjen_aids_2023}, 
Magic \cite{uci_magic_2007}, 
Clients \cite{uci_credit_2008}, 
SmartGrid \cite{kaggle_smartgrid_2020}, 
Diabetic \cite{esfandiari2022label}, 
Avila \cite{uci_avila_2018}, 
DryBean, 
Flipkart, 
CreditScore (Kaggle repository), 
and Malware (UCI repository). 
The first five sets have binary labels, and 
the last five have multi-class labels. 
(Although WNN was designed for binary label, 
it naturally applies to multi-class label.)  
On each set, we randomly selected 50\% 
of the points for training and used the 
rest for testing. Features were standardized, 
and all results were averaged over 20 
random splits.

In addition to WNN, we also compare RWNNs 
with representative compression 
methods including subsampling, NOSC \cite{gottlieb2014near}, and SNC \cite{kusner2014stochastic}. 
Euclidean distance is used by all methods. 
All results, including computation time, were 
generated on a machine with an 8-core Intel 11th Generation Core i7 processor, 32 GB of RAM, 
and a clock speed of 2.50 GHz. Source code 
is at \footnote{\url{https://github.com/Shayansh9674/RWNN.git}}.

\subsection{Comparison to WNN}

Performance of WNN, RWNNs and RWNN are shown in Table \ref{tab:wnnvsrwnn}. Condensing rate is $|\tilde{S}|/|S|$, 
and RWNNs sets $\theta = |\tilde{S}|/|S|$ to achieve 
the same rate as others. The error standard deviations 
of WNN and RWNNs across different datasets are around 0.0012 and 0.0016, respectively.

\begin{table*}[t!]
\def\arraystretch{1.3}
\centering
\normalsize
\begin{tabular}{l|c|cc|ccc|ccc}\hline
\bf Dataset & \bf Size 
& \multicolumn{2}{c|}{\bf Condensing Rate} 
& \multicolumn{3}{c|}{\bf Classification Error} 
& \multicolumn{3}{c}{\bf Compression Time} \\ \cline{3-10}
& & WNN & RWNN 
& WNN & RWNNs & RWNN 
& WNN & RWNNs & RWNN \\ \hline
Aids        & 15k   & 0.32 & 0.45 & 0.337 & 0.343 & 0.341 & 420 & 0.798 & 0.814 \\
Magic       & 19k   & 0.16 & 0.26 & 0.163 & 0.178 & 0.172 & 4219 & 0.749 & 0.786 \\
Clients     & 30k   & 0.23 & 0.35 & 0.207 & 0.220 & 0.219 & 2230 & 3.121 & 3.225 \\
{\small SmartGrid}  & 60k   & 0.05 & 0.14 & 0.047 & 0.077 & 0.072 & 3828 & 7.692 & 7.909 \\
\hline

Avila       & 7k    & 0.25 & 0.34 & 0.277 & 0.298 & 0.282 & 46 & 0.151 & 0.153 \\
DryBean     & 8k    & 0.08 & 0.12 & 0.085 & 0.087 & 0.086 & 17 & 0.234 & 0.235 \\
Flipkart    & 30k   & 0.28 & 0.36 & 0.315 & 0.325 & 0.322 & 2702 & 1.488 & 1.569 \\
\hline
\end{tabular}
\caption{Performance of WNN, RWNNs and RWNN}
\label{tab:wnnvsrwnn}
\end{table*}

Compared to WNN, we see that both RWNN and RWNNs consume substantially less compression time, with 
100-1000 times speedup on most datasets. 
They also obtain similar classification 
accuracy to WNN, demonstrating a more optimal 
tradeoff between efficiency and accuracy. 

We also see that RWNN generates larger 
condensed sets, which is consistent with the 
intuition. This, however, does not appear to 
affect its efficiency, since its compression 
time is close to the time of RWNNs and remains 
significantly lower than the time of WNN. 

\subsection{Comparison to Other Methods}

\begin{figure}[t!]
\centering
\includegraphics[width=0.45\textwidth]{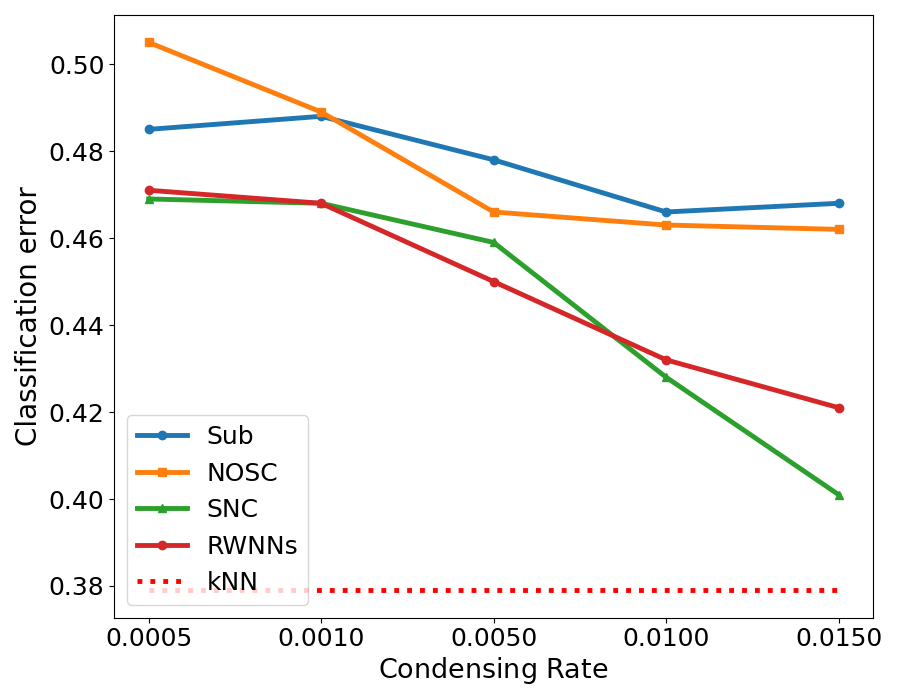}%
\vspace{-5pt}
\caption{Error versus Condensing
Rate on Diabetic}
\label{fig1:vsm}
\end{figure}

Table \ref{tab:comparison} shows the results 
of other representative compression methods. 
Condensing rate is set to 1\% (approximately 
for NOSC). 
To meet this goal, we stop the greedy 
search of WNN early and denote the resulting 
algorithm as WNNs. For NOSC, hyperparameter $\gamma$ indirectly determines the condensing rate, 
and we manually tuned it so the resulting 
rate is as close to 1\% as possible. Table \ref{tab:noscgamma} gives several example 
choices of $\gamma$ and their resulting 
condensing rates. Figure \ref{fig1:vsm} 
shows the performance across different 
condensing rates.

\begin{table*}[t!]
\setlength{\tabcolsep}{10pt}
\centering
\def\arraystretch{1.3}
\begin{tabular}{l|c|c|c|c|c|c|c}\hline
\bf Dataset & \bf Size & \bf 1NN & \bf Sub 
& \bf NOSC & \bf SNC & \bf WNNs & \bf RWNNs \\ \hline

Aids & 15k & 0.374 & 0.381 
& 0.407 & 0.362 & \underline{0.312} & \underline{0.337}\\

Magic & 19k & 0.192 & 0.302 
& 0.376 & 0.268 & \underline{0.194} & \underline{0.246} \\

Clients & 30k & 0.270 & 0.257 
& 0.262 & 0.275 & \underline{0.217} & \underline{0.225} \\ 

SmartGrid & 60k & 0.078 & 0.140 
& 0.153 & \underline{0.028} & \underline{0.057} & 0.110\\

Diabetic & 122k & 0.379 & 0.459 
& 0.530 & \underline{0.427} & - & \underline{0.432} \\ \hline 

Avila & 7k & 0.267 & 0.438 
& - & 0.507 & \underline{0.358}  & \underline{0.421}  \\ 

DryBean & 8k & 0.101 & 0.112  
& - & 0.194 & \underline{0.087}& \underline{0.101}\\

Flipkart & 30k &0.309& 0.482   
& - & \underline{0.411} & \underline{0.442} & 0.501\\ 

CreditScore & 100k & 0.253 & 0.389   
& - & \underline{0.305} & - & \underline{0.351}\\

Malware & 332k &  0.063 & 0.340   
& - &\underline{0.282} & - & \underline{0.220}\\ \hline
\end{tabular}
\caption{Classification Error. On each dataset, 
two lowest errors of the condensing algorithms 
are underlined. NOSC was designed 
for binary class and does not apply 
to the bottom five multiclass datasets. 
WNNs did not terminate in a reasonable time 
on Diabetic, CreditScore and Malware.}
\label{tab:comparison}
\end{table*}

\begin{table}[t!]
\def\arraystretch{1}
\centering
\begin{tabular}{ccc}\hline
\bf Method & \bf Compression Time  
& \bf Inference Time  \\ \hline
1NN & 0 & 5.48\\ 
Sub & 0 & 0.14\\ 
NOSC & 0.23 & 0.13 \\ 
SNC & 51.95 & 0.15 \\ 
WNNs & 887.85 & 0.92\\ 
RWNNs & 7.83 &  0.91 \\\hline 
\end{tabular}
\caption{Time (sec) on SmartGrid}
\label{tab:time}
\end{table}

\begin{table}[t!]
\def\arraystretch{1.2}
\centering
\begin{tabular}{c|c|c|c|c|c}\hline
$\gamma$ & 13.5 & 11 & 7.8 & 6.9 & 6.4 \\ \hline 
NOSC Rate & .06e-2 & .11e-2 & .53e-2 & 1e-2 & 1.6e-2 \\ \hline 
Target Rate & .05e-2 & .10e-2 & .50e-2 & 1e-2 & 1.5e-2 \\ \hline 
\end{tabular}
\caption{Example $\gamma$ (top row) 
and resulting condensing rate 
(bottom rows) for NOSC on Diabetic}
\label{tab:noscgamma}
\end{table}

In Table \ref{tab:comparison}, 
we observe that WNNs often achieves 
the lowest error, followed by RWNNs. 
This demonstrates the strength of 
this family of compression methods 
under low condensing rate. 
SNC also performs well such as 
on SmartGrid and CreditScore, 
but its compressed sets are synthetic 
and thus may have limited applicability 
due to the lack of interpretability. 
Besides, SNC compression is also 
more computationally expensive 
than RWNN, though less than WNN. 
Table \ref{tab:time} shows its 
compression time on SmartGrid. 
(WNNs/RWNNs takes more 
inference time because they compute 
weighted distance, while other methods 
compute unweighted distance.) 
Overall, the results 
suggest that RWNNs achieves a better 
efficiency-accuracy tradeoff.

Finally, Table \ref{tab:add_metrics} shows 
additional results under other performance 
metrics, under the same setting as in Table 
\ref{tab:wnnvsrwnn}. It also shows that 
RWNN and RWNNs perform closely to WNN.

\begin{table}[t!]
\def\arraystretch{1.2}
\centering
\begin{tabular}{ccccc}\hline
\bf Dataset & \bf Method & \bf Precision & \bf Recall & \bf F1 \\ \hline
\multirow{3}{*}{Aids} 
& WNN   & 0.441 & 0.266 & 0.332 \\
& RWNN  & 0.409 & 0.290 & 0.339 \\
& RWNNs & 0.416 & 0.284 & 0.337 \\ \hline
\multirow{3}{*}{Avila} 
& WNN   & 0.724 & 0.643 & 0.668 \\
& RWNN  & 0.714 & 0.645 & 0.667 \\
& RWNNs & 0.688 & 0.602 & 0.625 \\\hline
\end{tabular}
\caption{Performance under other Metrics}
\label{tab:add_metrics}
\end{table}

\section{Conclusion}

This study accelerates a theoretically 
elegant data compression algorithm WNN 
for nearest neighbor classification. 
The proposed algorithm RWNN simply 
randomizes the greedy ball search 
process of WNN, which substantially 
reduces compression time of WNN 
while maintaining its classification 
accuracy and certain theoretical properties 
including consistency and generalization 
bound. We further propose a heuristic 
variant of RWNN, named RWNNs, which allows 
one to specify the condensing rate by early 
stopping the (randomized) ball search process. 
Experimental results on multiple datasets 
demonstrate the efficacy of RWNN and RWNNs. 

\section{Acknowledgment}

Claude was used to proofread the 
paper and refine the proof of Remark \ref{rem:rwnn}.

\bibliographystyle{plain}
\bibliography{reference}

@article{Hart1968CNN,
  author  = {Hart, Peter E.},
  title   = {The Condensed Nearest Neighbor Rule},
  journal = {IEEE Transactions on Information Theory},
  volume  = {14},
  number  = {3},
  pages   = {515--516},
  year    = {1968},
  doi     = {10.1109/TIT.1968.1054155}
}

@article{Gates1972,
  author  = {Gates, G. W.},
  title   = {The Reduced Nearest Neighbor Rule},
  journal = {IEEE Transactions on Information Theory},
  volume  = {18},
  number  = {3},
  pages   = {431--433},
  year    = {1972},
  doi     = {10.1109/TIT.1972.1054809}
}

@inproceedings{Angiulli2005,
  author    = {Angiulli, Fabrizio},
  title     = {Fast Condensed Nearest Neighbor Rule},
  booktitle = {Proceedings of the 22nd International Conference on Machine Learning (ICML)},
  pages     = {25--32},
  year      = {2005},
  publisher = {ACM},
  doi       = {10.1145/1102351.1102355}
}

@article{chen2018explaining,
  title={Explaining the success of nearest neighbor methods in prediction},
  author={Chen, George H and Shah, Devavrat and others},
  journal={Foundations and Trends{\textregistered} in Machine Learning},
  volume={10},
  number={5-6},
  pages={337--588},
  year={2018},
  publisher={Now Publishers, Inc.}
}

@inproceedings{kusner2014stochastic,
  title={Stochastic neighbor compression},
  author={Kusner, Matt and Tyree, Stephen and Weinberger, Kilian and Agrawal, Kunal},
  booktitle={International Conference on Machine Learning},
  pages={622--630},
  year={2014},
  organization={PMLR}
}

@article{gottlieb2014near,
  title={Near-optimal sample compression for nearest neighbors},
  author={Gottlieb, Lee-Ad and Kontorovich, Aryeh and Nisnevitch, Pinhas},
  journal={Advances in Neural Information Processing Systems},
  volume={27},
  year={2014}
}

@inproceedings{esfandiari2022label,
  title={Label differential privacy via clustering},
  author={Esfandiari, Hossein and Mirrokni, Vahab and Syed, Umar and Vassilvitskii, Sergei},
  booktitle={International Conference on Artificial Intelligence and Statistics},
  pages={7055--7075},
  year={2022},
  organization={PMLR}
}

@article{CoverHart1967,
  author  = {Cover, Thomas M. and Hart, Peter E.},
  title   = {Nearest Neighbor Pattern Classification},
  journal = {IEEE Transactions on Information Theory},
  volume  = {13},
  number  = {1},
  pages   = {21--27},
  year    = {1967},
  doi     = {10.1109/TIT.1967.1053964}
}

@inproceedings{Gottlieb2024,
  author    = {Gottlieb, Lee-Ad and Sharabi, Tomer and Weiss, Roni},
  title     = {Weighted Distance Nearest Neighbor Condensing},
  booktitle = {Proceedings of the 41st International Conference on Machine Learning},
  year      = {2024},
  pages     = {16153--16166},
  publisher = {PMLR}
}

@misc{tanshihjen_aids_2023,
  author       = {Tan, Shih-Jen},
  title        = {AIDS Clinical Trials Dataset},
  year         = {2023},
  howpublished = {\url{https://www.kaggle.com/datasets/tanshihjen/aids-clinical-trials}},
  note         = {Accessed: 2025-06-25}
}

@misc{uci_magic_2007,
  author       = {{UCI Machine Learning Repository}},
  title        = {MAGIC Gamma Telescope Data Set},
  year         = {2007},
  howpublished = {\url{https://archive.ics.uci.edu/dataset/159/magic+gamma+telescope}},
  note         = {Accessed: 2025-06-25}
}

@misc{uci_credit_2008,
  author       = {Yeh, I-Cheng and Lien, Che-hui},
  title        = {Default of Credit Card Clients Dataset},
  year         = {2008},
  howpublished = {\url{https://archive.ics.uci.edu/dataset/350/default+of+credit+card+clients}},
  note         = {Accessed: 2025-06-25}
}

@misc{kaggle_smartgrid_2020,
  author       = {Breviglieri, P. C.},
  title        = {Smart Grid Stability Dataset},
  year         = {2020},
  howpublished = {\url{https://www.kaggle.com/datasets/pcbreviglieri/smart-grid-stability}},
  note         = {Accessed: 2025-06-25}
}

@misc{uci_avila_2018,
  author       = {Dua, Dheeru and Graff, Casey},
  title        = {Avila Dataset},
  year         = {2018},
  howpublished = {\url{https://archive.ics.uci.edu/dataset/459/avila}},
  note         = {Accessed: 2025-06-25}
}

\end{document}